\pdfoutput=1
\documentclass[11pt]{article}
\usepackage[preprint]{acl}
\usepackage{times}
\usepackage{latexsym}
\usepackage[T1]{fontenc}
\usepackage[utf8]{inputenc}
\usepackage{microtype}
\usepackage{inconsolata}
\usepackage{graphicx}
\usepackage{xspace}
\usepackage{booktabs}
\usepackage{multirow}
\usepackage{colortbl}
\usepackage{adjustbox}
\usepackage{amsmath}
\usepackage{amsfonts}
\usepackage{amssymb}
\usepackage{algorithm}
\usepackage{algorithmic}
\usepackage[many]{tcolorbox}
\newcommand{\themodel}{STRIVE\xspace} 

\title{\themodel: Structured Reasoning for Self-Improvement in \\Claim Verification}
\author{
Haisong Gong \textsuperscript{1,2},
Jing Li,
Junfei Wu \textsuperscript{1,2},
Qiang Liu \textsuperscript{1,2},
Shu Wu \textsuperscript{1,2},
Liang Wang \textsuperscript{1,2}
\\
\\
\textsuperscript{1}New Laboratory of Pattern Recognition (NLPR)\\Institute of Automation, Chinese Academy of Scieneces\\
\textsuperscript{2}School of Artificial Intelligence, University of Chinese Academy of Sciences\\
gonghaisong2021@ia.ac.cn,\ junfei.wu@cripac.ia.ac.cn,\ \{qiang.liu, shu.wu, wangliang\}@nlpr.ia.ac.cn
}

\begin{document}
\maketitle
\begin{abstract}
Claim verification is the task of determining whether a claim is supported or refuted by evidence. Self-improvement methods, where reasoning chains are generated and those leading to correct results are selected for training, have succeeded in tasks like mathematical problem solving. However, in claim verification, this approach struggles. Low-quality reasoning chains may falsely match binary truth labels, introducing faulty reasoning into the self-improvement process and ultimately degrading performance. To address this, we propose \themodel: Structured Reasoning for Self-Improved Verification. Our method introduces a structured reasoning design with Claim Decomposition, Entity Analysis, and Evidence Grounding Verification. These components improve reasoning quality, reduce errors, and provide additional supervision signals for self-improvement. \themodel begins with a warm-up phase, where the base model is fine-tuned on a small number of annotated examples to learn the structured reasoning design. It is then applied to generate reasoning chains for all training examples, selecting only those that are correct and structurally sound for subsequent self-improvement training. We demonstrate that \themodel achieves significant improvements over baseline models, with a 31.4\% performance gain over the base model and 20.7\% over Chain of Thought on the HOVER datasets, highlighting its effectiveness.

\end{abstract}
\section{Introduction}

\begin{figure}[tb]
  \includegraphics[width=0.99\columnwidth]{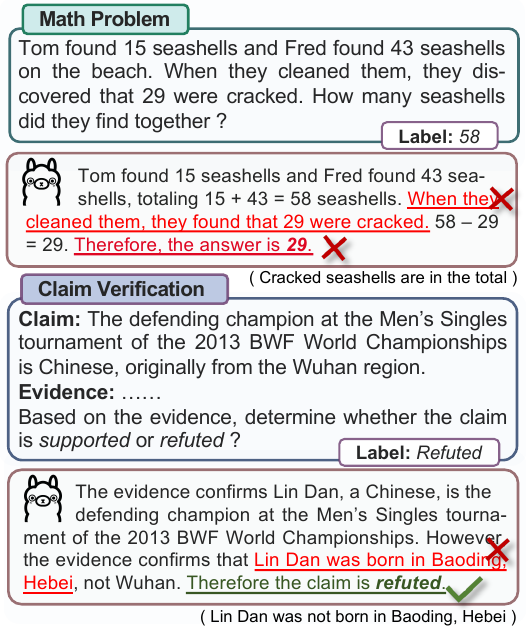}
  \caption{In math problem-solving, incorrect reasoning chains lead to mismatched final answers, while in claim verification, incorrect reasoning can still match the binary truth labels. Evidence omitted in figure.}
  \label{fig:intro_fig}
\end{figure}
The proliferation of misinformation is a major challenge in today's society, eroding trust in digital information and affecting domains such as public health~\cite{Naeem2020TheC} and politics~\cite{mishra2022factify}. As a result, claim verification---determining whether a claim is supported or refuted by evidence---has become crucial for ensuring reliable information.

Claim verification is a natural language inference task. While large language models (LLMs) demonstrate impressive reasoning capabilities, they struggle with claims requiring multi-hop evidence synthesis. Methods like few-shot learning and chain-of-thought (CoT) prompting have not yielded satisfactory results for complex claims.

Self-improvement methods have been successfully applied in domains like mathematical problem solving and commonsense reasoning~\cite{zelikman2022star,hosseini2024v}. These methods involve generating reasoning chains and answers, selecting the ones that lead to correct results, and using them to train the model for improved performance. However, directly applying self-improvement to claim verification proves ineffective. In mathematical problem solving and commonsense reasoning, the answers are typically numerical values or short phrases, with an almost infinite range of possibilities. This makes it difficult for incorrect reasoning chains to match the correct answer. In contrast, in claim verification, incorrect reasoning chains can easily match the binary truth labels (\textit{Supported}/\textit{Refuted}), leading to their selection for training, as illustrated in Figure~\ref{fig:intro_fig}. This introduces low-quality data into the training process, which undermines model performance. Our experiments validate this, showing that naive self-improvement degrades performance in claim verification task, with performance dropping by 4.5\% and 1.1\% on HOVER-2 and HOVER-4 (Table \ref{tab:main_results}), respectively.

Upon closer examination, we found that reasoning chains in claim verification often suffer from issues such as evidence confusion, entity misidentification, and omission of key information---even when the final truthfulness decision is correct. To address this, we propose a structured reasoning design. Our design requires three key components: 1) Claim Decomposition: Breaking complex claims into smaller, manageable subclaims; 2) Entity Analysis: Linking ambiguous terms to grounded entities; and 3) Evidence Grounding Verification: Citing specific evidence snippets at each reasoning step. This structure improves the quality of reasoning chains and reduces the chance of error-prone chains being included in the training data. Moreover, the structural constraints provide additional supervision signals that, when combined with binary labels, enable more effective filtering for self-improvement.

In this work, we introduce \themodel: Structured Reasoning for Self-Improved Verification. \themodel begins with a warm-up phase, where a base model is fine-tuned to generate structured reasoning chains using a small set of annotated examples. This prepares the model to follow the desired reasoning format. Next, the warm-up model is used to generate reasoning chains and verification results for all examples in the training set. We select reasoning chains that both lead to correct verification results and follow the correct structural format. Finally, the correct, structured reasoning chains are used as training data to fine-tune the base model. Our experiments demonstrate that, compared to the base model, \themodel achieves an average improvement of 31.4\% and 20.7\% over standard Chain of Thought (CoT) reasoning on HOVER datasets, underscoring the effectiveness of our approach.

Our contributions are summarized as follows:
\begin{itemize}
    \item We identify the limitations of freely generated reasoning chains in claim verification for self-improvement and propose \themodel, a solution that combines structured reasoning to address these challenges.
    \item We design a structured reasoning approach incorporating Claim Decomposition, Entity Analysis, and Evidence Grounding Verification, enhancing reasoning quality.
    \item Our experiments show that \themodel achieves significant improvements over the baseline model across multiple datasets, demonstrating the effectiveness of our approach.
\end{itemize}

\section{Related Works}
\subsection{Claim Verification}

Early approaches to claim verification focused on fine-tuning pre-trained models, either by concatenating evidence and claims into a single input~\cite{aly2021feverous, thorne2018fever, hu2022dual} or processing evidence separately and aggregating the results~\cite{soleimani2020bert, jiang2021exploring,gi-etal-2021-verdict}. Graph Neural Networks have also been applied to capture relationships between evidence pieces~\cite{gong2024heterogeneous, Zhaotransxh, Chenevidencenet}. With the impressive generative capabilities demonstrated by large language models (LLMs), many studies have turned to LLMs for claim verification~\cite{ma-etal-2024-ex}. FACT-GPT~\cite{factgpt} and FactLlama~\cite{cheung2023factllama} fine-tune LLMs to directly predict the truthfulness of claims. Factscore~\cite{min2023factscore} employs systematic decomposition to assess the factuality of individual claim segments, while ProgramFC~\cite{pan2023fact} frames claim verification as step-wise program execution. Other works, such as \citet{li2023self}, \citet{chen2022generating}, and \citet{rani2023factify}, transform the verification task into a series of sub-questions to be answered.

\subsection{Chain of Thought Reasoning (CoT)}
Chain of Thought (CoT) reasoning~\cite{wei2022chain} was proposed to help LLMs solve complex problems by breaking them down into intermediate step-by-step reasoning. \citet{kojima2022large} demonstrated that adding a prompt such as ``Let's think step by step'' significantly boosts LLM performance. CoT reasoning has been applied to a variety of tasks, including claim verification. Studies like \citet{hu2023large} and \citet{dougrez2024assessing} evaluate CoT methods in different contexts. FOLK~\cite{wang2023explainable} leverages LLMs to transform claims into first-order logic as intermediate steps for explainable verification.

\subsection{Self-Improvement Methods}
Self-improvement methods for LLMs have garnered attention in recent years, where models are fine-tuned on their self-generated solutions, optionally iterating this process~\cite{hosseini2024v}. $\text{ReST}^{EM}$~\cite{singh2023beyond} generates reasoning chains for solving math problems and selects those leading to correct answers for retraining. RFT~\cite{yuan2023scaling} enhances reasoning chain diversity by sampling multiple chains before selection. STaR~\cite{zelikman2022star} introduces hints during reasoning generation for chains that lead to incorrect results. V-STaR~\cite{hosseini2024v} incorporates Direct Preference Optimization~\cite{rafailov2023direct} into the self-improvement process. Our method shares similarities with STaR. We are the first to apply self-improvement to claim verification. We also highlight the unique challenges of claim verification, distinguishing it from tasks like math problem-solving, and address these challenges through the integration of structured reasoning design.

\newtcolorbox{mybox}{
    colframe=gray!50,    
    colback=white,       
    coltitle=black,      
    boxrule=0.8mm,       
    arc=2mm,             
    top=1mm,             
    bottom=1mm,          
    left=2mm,            
    right=2mm,           
    before skip=2mm,     
    after skip=2mm,      
}
\definecolor{sub}{HTML}{f6f6f6}

\section{Method}
In this section, we introduce our approach to claim verification. We first define the task itself, followed by our Structured Reasoning Design, which is specifically tailored for claim verification. Finally, we describe how this structure is applied within the self-improvement process to enhance the verification model’s performance.

\subsection{Task Formulation}
The task of claim verification is to determine the truthfulness of a given claim based on a set of evidence. Our approach aims to generate reasoning chains connecting the claim and evidence to a final prediction, rather than outputting the answer directly. However, it’s important to note that reasoning chains are not provided in the verification datasets, and only the final truth label is available.

Formally, given a claim $c$ and an evidence set $\mathcal{E} = \{e_1, e_2, \dots, e_n\}$, where each $e_i$ is a descriptive piece of evidence (such as a sentence or paragraph), the goal is to obtain a model that can generate a reasoning chain $r$ that forms the intermediate reasoning steps. The final prediction $\hat p$ is made based on the reasoning chain, where $\hat p \in \{\textit{Supported}, \textit{Refuted}\}$.

\begin{figure}[tb]
  \includegraphics[width=0.99\columnwidth]{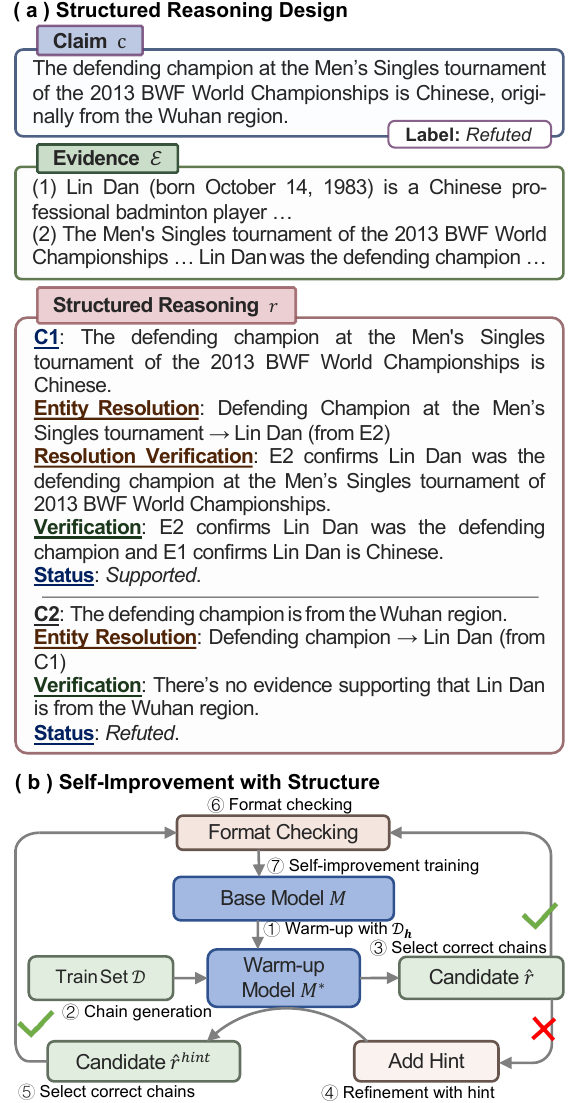}
  \caption{(a) An example demonstrating our Structured Reasoning Design. (b) Overview of \themodel.}
  \label{fig:main}
\end{figure}

\subsection{Structured Reasoning Design}\label{sec:structure}
Freely generated reasoning chains often struggle with verifying complex claims due to issues like evidence confusion, entity misidentification, and omission of key information. These problems can lead to low-quality chains being selected during self-improvement, affecting performance.

To address this, we introduce a Structured Reasoning Design that guides the verification process systematically. Our Structured Reasoning Design consists of three key components: Claim Decomposition, Entity Analysis, and Evidence Grounding Verification. An example is illustrated in Figure~\ref{fig:main}.

\subsubsection{Claim Decomposition}
Complex claims typically describe multiple facets of a situation, such as in our example (Figure~\ref{fig:main}), where the claim covers the champion's nationality, victory, and origin. When a claim involves multiple elements, decomposing it into subclaims allows for independent verification of each component, a method known to improve performance in claim verification~\cite{gong2024navigating,min2023factscore}. In our design, we structure the reasoning chain into distinct blocks, each dedicated to a specific subclaim, labeled as \underline{\texttt{C1:}}, \underline{\texttt{C2:}}, and so on. This block-based structure helps mitigate the risk of overlooking key aspects during verification, which is common in free reasoning. Each subclaim block concludes with a verification result, labeled as \underline{\texttt{Status:}}, to make the reasoning easy to follow and parse.
\subsubsection{Entity Analysis}
Complex claims often involve ambiguous or unspecified entities or pronouns that may appear after Claim Decomposition. To resolve these ambiguities, we designed a two-step entity analysis process: Entity Resolution and Resolution Verification.
\begin{itemize}
    \item \textbf{Entity Resolution} aims to identify the specific entity corresponding to an ambiguous term by leveraging evidence, labeled as \underline{\texttt{Entity Resolution:}} in the reasoning chain. For instance, the term ``defending champion'' in the example claim (Figure~\ref{fig:main}) is resolved to the specific person ``Lin Dan'' using information from the evidence.
    \item \textbf{Resolution Verification} ensures the correctness of the Entity Resolution, labeled as \underline{\texttt{Resolution Verification:}} in the reasoning chain. While Entity Resolution may correctly match an ambiguous term to a known entity in many cases, errors can occur. For example, if the claim stated a ``Doubles tournament'' instead of a ``Singles tournament'' in the example, the Entity Resolution step might still match the term ``defending champion at the Men's Doubles tournament'' to ``Lin Dan'', which is incorrect for the revised claim. Resolution Verification cross-checks the entity to ensure that the resolution is accurate, preventing errors in the reasoning process.
\end{itemize}

This two-step design ensures more precise entity analysis, minimizing the risk of errors that could impact the overall verification. We do not always require both steps to appear simultaneously. For instance, in example C2, the Entity Resolution is drawn from C1, and therefore the Resolution Verification step is omitted.

\subsubsection{Evidence Grounding Verification}
Our structured reasoning framework incorporates explicit verification steps at multiple stages. In addition to the Resolution Verification process mentioned earlier, each subclaim undergoes a final verification step, labeled as \underline{\texttt{Verification:}}. This step evaluates the subclaim as a whole, considering the resolved entities and assessing its validity against the provided evidence. It also serves as an explanation for the truthfulness decision of the subclaim.

Our structured reasoning format enforces grounding at every stage of the reasoning chain. As illustrated in Figure~\ref{fig:main}, both the resolution and verification processes explicitly cite the corresponding evidence or subclaim identifiers (e.g., C1, E2). This ensures that conclusions are drawn from verifiable sources, making the reasoning more transparent and easier for humans to interpret and verify.

\subsection{Self-Improvement with Structure}
In this section, we describe how \themodel leverages the Structured Reasoning Design to improve model's performance in claim verification,  with the general flow of this process shown in the bottom part of Figure~\ref{fig:main}. Given a training set \begin{equation*}\mathcal{D} = \{(c_1, \mathcal{E}_1, p_1), \cdots, (c_N, \mathcal{E}_N, p_N)\},\end{equation*} where each $c_i$ is a claim, $\mathcal{E}_i$ is the corresponding evidence set, and $p_i$ is the final label, \themodel follows three main steps to complete the self-improvement process: Structured Warm-up, Reasoning Chain Generation and Selection, and Self-Improvement Training.
\subsubsection{Structured Warm-up}\label{sec:ourprompt}
Given a base model $M$, the purpose of the warm-up phase is to fine-tune $M$ into $M^*$, enabling it to generate reasoning chains conforming to our predefined structure. We use a preset prompt template $T(c,\mathcal{E})$ as follows:

\begin{mybox}
\textit{Based on the evidence, determine if the claim is supported by the evidence or refuted by it. Output the reasoning chain.\\
Claim: \texttt{[claim text $c$]}\\
Evidence: \texttt{(1)[evidence text $e_1$](2)...}}
\end{mybox}
Since reasoning chains are not available in the dataset, we manually annotate a very small set \begin{equation*} \mathcal{D}_h=\{(c_1^h,\mathcal{E}^h_1,r^h_1,p^h_1),\cdots,(c_H^h,\mathcal{E}^h_H,r^h_H,p^h_H)\}\end{equation*} with $H$ examples, where each pair $(c^h_i,\mathcal{E}^h_i)$ is annotated with a reasoning chain $r_i^h$ that following our prescribed structure. We then fine-tune the base model $M$ on $\mathcal{D}_h$ using $T(c^h_i,\mathcal{E}_i^h)$ as input and $r_i^h$ as output, resulting in a new model $M^*$. Notably, only a very small number of examples is needed for this fine-tuning, since our structure is guided by keywords (e.g., ``Verification:'') that activate the model’s inherent reasoning capabilities.
\subsubsection{Reasoning Chain Generation and Selection}
With the fine-tuned model $M^*$, our next objective is to leverage $M^*$ and training set $\mathcal{D}$ to generate and select high-quality structured reasoning chains for further self-improvement. Inspired by the chain generation strategy in STaR~\cite{zelikman2022star}, our method involves three stages: (i) Initial Generation and Selection and (ii) Refinement with Hint and (iii) Format Checking:

\begin{algorithm}[t]
\caption{\themodel}
\label{alg:STRIVE}
\textbf{Input:} Base model $M$, training set $\mathcal{D}$, annotated set $\mathcal{D}_h$\\
\textbf{Output:} Improved model $M_{st}$

\begin{algorithmic}[1]
\STATE $M^* \gets \text{train}(M,\mathcal{D}_h)$  \textcolor{gray}{// Structured Warm-up}
\STATE $\hat{r}_i \gets M^*(T(c_i, \mathcal{E}_i)),\ \hat{p}_i \gets J(\hat{r}_i)\ \forall i \in [1, N]$  \textcolor{gray}{//Get reasoning chains and predicted labels}
\STATE $\mathcal{D}_1 \gets \{(c_i, \mathcal{E}_i, \hat{r}_i) \mid \hat{p}_i = p_i \}$ \textcolor{gray}{//Select correct reasoning chains based on ground truth label}
\STATE  $\hat{r}_i^{hint} \hspace{-1mm}\gets \hspace{-1mm} M^*(T^{hint}(c_i,\mathcal{E}_i)),\ \hat{p}_i^{hint}\hspace{-1mm} \gets \hspace{-1mm} J(\hat{r}_i^{hint}),$\\ $\forall i\in[1,N],\text{ where }\hat{p}_i \neq p_i$ \textcolor{gray}{//Regenerate reasoning chains for incorrect predictions}
\STATE $\mathcal{D}_2 \gets \{(c_i, \mathcal{E}_i, \hat{r}_i^{hint}) \mid \hat p_i\neq p_i,\hat{p}_i^{hint} = p_i \}$\textcolor{gray}{//Select corrected reasoning chains}
\STATE $\mathcal{D}_{st} \gets \{(c_i, \mathcal{E}_i, r_i) \mid r_i \in (\mathcal{D}_1 \cup \mathcal{D}_2), f(r_i) = \text{True}\}$ \textcolor{gray}{//Select structurally valid chains}
\STATE $M_{st}\gets \text{train}(M,\mathcal{D}_{st}\cup\mathcal{D}_h)$ \textcolor{gray}{//Final Self-improvement training}
\end{algorithmic}
\end{algorithm}

\textbf{Initial Generation and Selection}: For each sample $(c_i,\mathcal{E}_i,p_i)\in \mathcal{D}$, we first generate a reasoning chain $\hat r_i$ using the prompt template $T(c_i,\mathcal{E}_i)$ with the model $M^*$:\begin{equation*}\hat{r_i}=M^*(T(c_i,\mathcal{E}_i))\quad \forall i\in[1,N].\end{equation*} A predicted label $\hat p_i$ is then derived from $\hat r_i$ by a rule-based function $J$: \begin{equation*}\hat p_i=J(\hat r_i),
\end{equation*}where if any subclaim in $\hat r_i$ is judged as \textit{Refuted}, then $\hat p_i=\textit{Refuted}$, and only if all subclaims are \textit{Supported}, then $\hat p_i=\textit{Supported}$. We select samples for which $\hat p_i$ matches the ground truth $p_i$ to form the set:\begin{equation*}\mathcal{D}_1=\{(c_i,\mathcal{E}_i,\hat r_i)\ |\ \hat p_i=p_i,\forall i\in [1,N]\}.\end{equation*}

\textbf{Refinement with Hint}: For samples where $\hat p_i\neq p_i$, we incorporate additional guidance by modifying the prompt. If $p_i = \textit{Supported}$, we add the hint: ``every detail in this claim is supported''; if $p_i=\textit{Refuted}$, we add ``the claim should be refuted, locate the error in the reasoning chain''. The modified prompt, inclusive of the hint, can be found in Appendix~\ref{sec:prompthint}. Using this modified prompt, we regenerate the reasoning chain: \begin{equation*}\begin{aligned}\hat r_i^{hint}&=M^*(T^{hint}(c_i,\mathcal{E}_i)),\\ \hat p_i^{hint}=J(\hat r_i^{hint})\quad &\forall i\in[1,N], \text{ where }\hat p_i\neq p_i.\end{aligned}\end{equation*}
Similarly, we select samples for which $\hat p_i^{hint}$ matches the ground truth $p_i$ to form the set:\begin{multline*}\mathcal{D}_2=\{(c_i,\mathcal{E}_i,\hat r_i^{hint})\ |\ \hat p_i\neq p_i,\hat p^{hint}_i= p_i,\\\forall i\in [1,N]\}\end{multline*} 

\textbf{Format Checking}: At this stage, both $\mathcal{D}_1$ and $\mathcal{D}_2$ contain reasoning chains that yield correct final predictions. To further ensure the quality of these reasoning chains, we apply a rule-based structural verification function $f$, where $f(\hat r_i)=\textit{True} \text{ or } \textit{False}$, indicating whether $\hat r_i$ follow our predefined structure. Specifically, we enforce three criteria:  
\begin{itemize}
\item Proper segmentation of subclaims, ensuring each reasoning step is explicitly delineated with a corresponding verification result.  
\item Correct evidence grounding, preventing incorrect references (e.g., citing non-existent evidence, citing C2 while verifying C1).  
\item Adherence to the structured format, ensuring reasoning steps are guided by predefined keywords (e.g. Entity Resolution:)
\end{itemize}
By applying this structural verification, we obtain a final set of reasoning chains $\mathcal{D}_{st}$, which not only yield correct conclusions but also adhere to the required structural format. Formally, it can be described as:
\begin{equation*}
\mathcal{D}_{st} = \{ (c_i, \mathcal{E}_i, r_i) \ |\ r_i \in (\mathcal{D}_1 \cup \mathcal{D}_2), f(r_i) = \textit{True} \}.
\end{equation*}

\subsubsection{Self-Improvement Training}
Finally, we fine-tune the base model $M$ using both the previously selected dataset $\mathcal{D}_{st}$ and the human-annotated dataset $\mathcal{D}_h$. Note that $M^*$ is discarded after generating the reasoning chains. This allows us to obtain the final model $M_{st}$ with the enhanced reasoning ability defined by our structured approach. The entire process is summarized in Algorithm \ref{alg:STRIVE}.

\definecolor{lightgray}{gray}{0.9}

\section{Experiment}

\begin{table*}[t]
  \centering
  \begin{adjustbox}{max width=\textwidth}
  \setlength{\tabcolsep}{3.5pt}
    \begin{tabular}{llcccc}
    \toprule
    \textbf{Model Family} & \textbf{Approach} & \textbf{HOVER-2}  & \textbf{HOVER-3}  & \textbf{HOVER-4}  & \textbf{FEVEROUS-S} \\
    \midrule
    \multirow{4}{*}{PT/FT Models} 
    & BERT-FC & 53.40  & 50.90  & 50.86 & 74.71 \\
    & LisT5 & 56.15 & 53.76 & 51.67 & 77.88 \\
    & RoBERTa-NLI & \underline{74.62} & 62.23 & 57.98 & 88.28 \\
    & MULTIVERS & 68.86 & 59.87 & 55.67 & 86.03 \\
    \midrule
    
    \multirow{4}{*}{Llama-3-8B} 
    & Zero-shot & 55.10  & 55.09 & 53.51 & 78.21 \\
    & Zero-shot + CoT & 63.76 & 57.13 & 57.47 & 84.94 \\
    & Few-shot & 55.33 & 55.63 & 52.86 & 79.17 \\
    & Few-shot + Structured CoT & 69.71 & \underline{66.71} & 59.63 & 85.67 \\
    \midrule
    
    \multirow{3}{*}{LoRA-Llama-3} 
    & LoRA Fine-tuning & 64.21 & 60.35 & \underline{60.34} & \underline{91.52} \\
    & STaR* (Self-Improvement) & 60.90  & 58.61 & 56.86 & 87.45 \\
    & \cellcolor{gray!20}\textbf{\themodel (Ours)} & \cellcolor{gray!20}\textbf{76.13} \scriptsize{$\pm$0.84} & \cellcolor{gray!20}\textbf{70.50} \scriptsize{$\pm$0.55} & \cellcolor{gray!20}\textbf{68.50} \scriptsize{$\pm$1.27} & \cellcolor{gray!20}\textbf{91.91} \scriptsize{$\pm$0.44} \\
    \bottomrule
    \end{tabular}%
    \end{adjustbox}
    \caption{Macro-F1 scores for claim verification models on HOVER and FEVEROUS-S datasets. ``PT/FT'' refers to pretrained/fine-tuned models.}
  \label{tab:main_results}%
\end{table*}%

\subsection{Datasets}
We evaluate \themodel using two publicly available datasets, following prior work in claim verification~\cite{gong2024navigating}. All evaluations are performed on the validation sets, since the test sets are not publicly released. Detailed information can be found in Appendix~\ref{sec:datasetinfo}. Results are reported using the Macro-F1 score.
\begin{itemize}
\item \textbf{HOVER}~\cite{jiang2020hover} \xspace This dataset comprises claims necessitating multi-step reasoning across multiple pieces of evidence and is categorized into three subsets: HOVER-2 for two-hop reasoning, HOVER-3 for three-hop reasoning, and HOVER-4 for four-hop reasoning.
\item \textbf{FEVEROUS-S}~\cite{aly2021feverous} \xspace Derived from the FEVEROUS dataset, this subset contains claims that rely solely on unstructured textual evidence. Compared to HOVER, claims in FEVEROUS-S generally exhibit lower complexity in reasoning.  
\end{itemize}

\subsection{Baselines}
To assess the effectiveness of \themodel, we compare it against a variety of baselines, including pretrained/fine-tuned models, prompt-based approaches using the base model, and base model fine-tuned on the training data.

For pretrained/fine-tuned models, we include the following: (i) BERT-FC~\cite{soleimani2020bert}: Pretrained BERT model~\cite{devlin2018bert} tailored for fact-checking tasks. (ii) LisT5~\cite{jiang2021exploring}: Pretrained T5 model~\cite{raffel2020exploring} specialized for fact-checking tasks. (iii) RoBERTa-NLI~\cite{nie2020adversarial}: Pretrained RoBERTa-large model~\cite{liu2019roberta} fine-tuned on four natural language inference datasets. (iv) MULTIVERS~\cite{wadden2022multivers}: A LongFormer model~\cite{beltagy2020longformer} fine-tuned on the FEVER~\cite{thorne2018fever} dataset.

For prompt-based approaches, we evaluate the following: (i) Zero-shot: The model predicts the final label directly, without requiring reasoning steps. (ii) Zero-shot + CoT: The model outputs reasoning chains before predicting the final label. (iii) Few-shot: Similar to zero-shot, but with the inclusion of labeled examples. (iv) Few-shot + Structured CoT: Similar to zero-shot + CoT, but with examples of structured reasoning chains from $\mathcal{D}_h$. All baseline prompts are similar to $T(c,\mathcal{E})$ for fairness, detailed in Appendix~\ref{sec:appendix_prompt_baseline}.

Finally, for base models that undergo fine-tuning, we compare with: (i) Lora Fine-tuning: Fine-tuning the base model on the training set $\mathcal{D}$ using only binary labels. (ii) STaR*~\cite{zelikman2022star}:  We re-implemented this self-improvement method for claim verification. While similar to our approach (Algorithm~ \ref{alg:STRIVE}), it differs in that reasoning chains are generated freely by the model, without structural constraints or a warm-up phase.

\subsection{Implementation Details}
We use Llama-3-8B-Instruct as the base model, as it is a widely used open-source language model. For all fine-tuning tasks, including those in baseline models and \themodel, we employ the GPU memory-efficient LoRA fine-tuning method~\cite{hu2021lora}, allowing our experiments to fit on a single consumer-grade GPU (e.g., NVIDIA 4090).

In the Structured Warm-up process, we use a small human-annotated dataset $\mathcal{D}_h$ containing only $H=10$ examples, with 8 labeled as \textit{Refuted} and 2 as \textit{Supported}. We prioritize teaching the model to identify errors rather than admitting correct claims. The intermediate model $M^*$ is obtained by fine-tuning on $\mathcal{D}_h$ for 10 epochs using LoRA. Despite the large number of epochs, only 0.1\% of the model's parameters are updated, ensuring the model’s overall performance is maintained while enforcing the prescribed structure. The training set $\mathcal{D}$ consists of $N=600$ examples, with reasoning chains generated at a temperature setting of 0.01. The final model $M_{st}$ is obtained by fine-tuning for 2 epochs on the union of the collected set and the human-annotated set, $\mathcal{D}_{st} \cup \mathcal{D}_h$.

\begin{figure}[t]
  \includegraphics[width=\columnwidth]{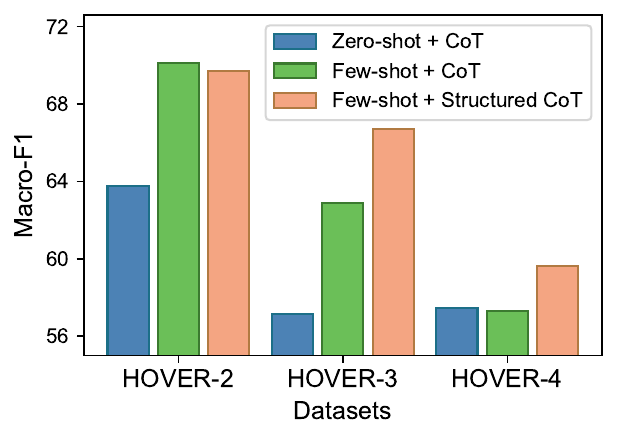}
  \caption{Comparison of Chain-of-Thought approaches with and without structured reasoning over HOVER datasets.}
  \label{fig:structure}
\end{figure}

\section{Results and Discussion}
\subsection{Overall Performance}
We present the overall results of \themodel and the baseline models in Table \ref{tab:main_results}. Among all methods, \themodel achieves the best performance, underscoring the effectiveness of our approach. The results also highlight the importance of explicitly reasoning steps in claim verification. Pretrained/fine-tuned models, zero-shot and few-shot methods that do not incorporate reasoning steps, generally perform worse than methods that explicitly utilize reasoning steps (e.g., Zero-shot + CoT, Few-shot + Structured CoT). 

We present reasoning chains generated by \themodel and \textbf{error analysis} in Appendix~\ref{sec:cases}.

\subsection{Effectiveness of Structured Reasoning}
In this section, we explain how Structured Reasoning contributes to performance improvement in claim verification.

\textbf{Structured Design Improves Reasoning Quality.}\xspace
We compare three approaches: Zero-shot + CoT, Few-shot + CoT, and Few-shot + Structured CoT. Zero-shot + CoT and Few-shot + CoT use chain-of-thought prompting without structural constraints. In Few-shot + CoT, reasoning examples are rewritten by removing structural elements from $\mathcal{D}_h$'s chains. In contrast, Few-shot + Structured CoT directly uses reasoning examples from $\mathcal{D}_h$ with the desired structure. As shown in Figure~\ref{fig:structure}, while Few-shot + CoT performs slightly better on simpler tasks like HOVER-2, structured reasoning shows significant advantages in more complex scenarios like HOVER-3 and HOVER-4. This confirms that structured reasoning improves reasoning quality, particularly for complex tasks.


\begin{figure}[t]
  \includegraphics[width=\columnwidth]{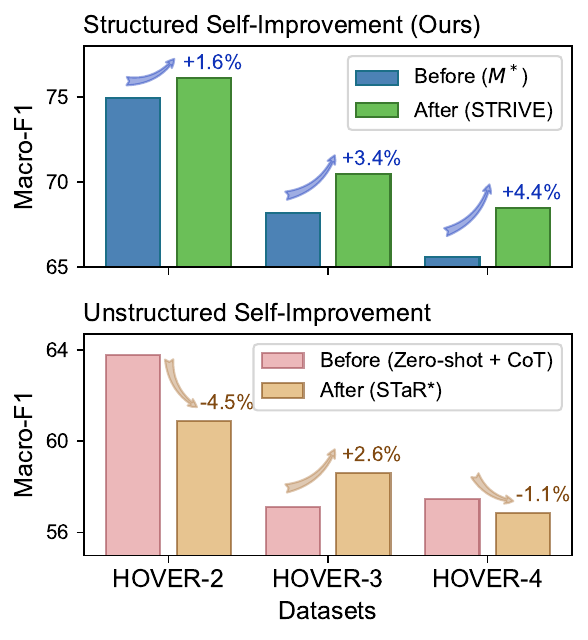}
  \caption{Comparison of self-improvement training results with (top) and without (bottom) structured reasoning over HOVER datasets.}
  \label{fig:improve}
\end{figure}

\textbf{Structured Design Helps in Self-Improvement Training.}\xspace 
Figure~\ref{fig:improve} compares self-improvement results with (top) and without (bottom) structured reasoning. In the top part, performance before self-improvement corresponds to the model $M^*$ after the Structured Warm-up phase. In the bottom part, ``before'' and ``after'' refer to Zero-shot + CoT and STaR*, respectively. The figure shows that structured reasoning leads to steady performance improvement after the warm-up model, with gains increasing as task complexity rises (1.6\% for 2-hop, 4.4\% for 4-hop). In contrast, without the structured design, improvements are inconsistent. As shown in the bottom of Figure~\ref{fig:improve}, performance even drops on HOVER-2 and HOVER-4 due to the inclusion of erroneous reasoning chains that incorrectly match the final label. These results highlight the importance of structured reasoning in ensuring successful self-improvement for claim verification.

\begin{table}[t]
  \centering
    \begin{adjustbox}{max width=\columnwidth}
  \setlength{\tabcolsep}{3.5pt}
    \begin{tabular}{lccc}
    \toprule
    \multicolumn{1}{c}{\textbf{Models}} & \textbf{HOVER-2} & \textbf{HOVER-3} & \textbf{HOVER-4} \\
    \midrule
    \multicolumn{1}{c}{STRIVE} & \textbf{76.13} & \underline{70.50}  & \textbf{68.50}\\
    \midrule
    $\text{STRIVE}_{\text{w/o CD}}$ & 73.69 & 67.36 & 64.03 \\
    $\text{STRIVE}_{\text{w/o EA}}$ & 75.96 & 67.36 & 64.77 \\
    $\text{STRIVE}_{\text{w/o EG}}$ & 75.08 & 69.42 & 68.22 \\
    \midrule
    $\text{\themodel}_{\text{w/o FC}}$ & \underline{76.11} & 69.69 & 66.89 \\
    $\text{\themodel}_{\text{w/o hint}}$ & 75.90  & 70.34 & 67.20 \\
    $\text{\themodel}_{2\text{ rounds}}$ & 75.98 & \textbf{70.55} & \underline{68.43} \\
    \bottomrule
    \end{tabular}%
    \end{adjustbox}
    \caption{Ablation study results. Macro-F1 scores for different model variants across the HOVER datasets. ``CD'' stands for Claim Decomposition, ``EA'' stands for Entity Analysis, ``EG'' stands for Evidence Grounding Verification, and ``FC'' stands for Format Checking.}
  \label{tab:ablate}
\end{table}

\subsection{Analysis of Structure Design}
In this section, we analyze the structural reasoning design of \themodel. As outlined in Section~\ref{sec:structure}, the core components---Claim Decomposition, Entity Analysis, and Evidence Grounding Verification---are introduced to address the issue of low-quality reasoning chains generated by the model. To assess the contribution of each component, we conduct an ablation study by removing each design element individually, resulting in three variants: $\text{STRIVE}_{\text{w/o CD}}$, $\text{STRIVE}_{\text{w/o EA}}$, and $\text{STRIVE}_{\text{w/o EG}}$. The results are presented in Table \ref{tab:ablate}.

From the table, it is evident that Claim Decomposition is the most impactful design; its removal causes performance to drop across all HOVER datasets. Entity Analysis also plays a significant role, particularly in more complex scenarios (HOVER-4), as complex claims often involve multiple ambiguous entities. In contrast, the Evidence Grounding Verification design has a smaller impact on performance but still contributes valuable benefits, primarily by improving human readability of the reasoning chains.

\subsection{Analysis of Self-Improvement Training}
To analyze the self-improvement training process, we propose three variants: $\text{\themodel}_{\text{w/o FC}}$, $\text{\themodel}_{\text{w/o hint}}$ and $\text{\themodel}_{2\text{ rounds}}$, with performance results shown in Table \ref{tab:ablate}.

(i) $\text{\themodel}_{\text{w/o FC}}$ removes the format checking step during the Reasoning Chain Selection process, causing a performance drop, particularly on the HOVER-4 dataset. (ii) $\text{\themodel}_{\text{w/o hint}}$ eliminates the hint-based refinement in the Reasoning Chain Generation process (steps 4 and 5 in Algorithm \ref{alg:STRIVE}), leading to a small performance decrease across the HOVER datasets. (iii) $\text{\themodel}_{2\text{ rounds}}$ adds an extra round of self-improvement training (repeating steps 2-7 in Algorithm \ref{alg:STRIVE} with $M^*$ replaced by $M_{st}$). However, this results in performance similar to the original model, indicating no significant benefit from the additional round. Therefore, we opt to use a single round of training.

Interestingly, unlike math problem solving~\cite{zelikman2022star,hosseini2024v}, where extra rounds are critical, \themodel performs well with a single round. This is likely due to its use of structured guidance, which activates the model's inherent reasoning abilities, and no new capabilities are gained from further training rounds.

\section{Conclusion}
We presented \themodel: Structured Reasoning for Self-Improved Verification, a method that integrates structured reasoning and self-improvement training for claim verification. By incorporating Claim Decomposition, Entity Analysis, and Evidence Grounding Verification, \themodel improves the quality of reasoning chains and enables more effective self-improvement. Our experiments demonstrate that \themodel significantly outperforms baseline approaches, highlighting the effectiveness of structured reasoning for self-improvement in claim verification. 

\section{Limitations}
Our approach relies on a structured warm-up phase that requires a small amount of annotated data. In our experiments, we selected 10 moderately difficult claims for reasoning chain labeling without further extensive sample filtering. While this approach has yielded positive results, we recognize that the choice of these samples may influence the subsequent model training. We believe that more carefully selected or diverse samples could further enhance the model's performance and provide additional insights into how sample selection impacts self-improvement. Additionally, our approach improves the claim verification capabilities of LLM in a resource-efficient manner. Both the quantity of annotations and the training strategies were designed for resource efficiency. This low-cost approach has proven effective for performance enhancement, but it also presents opportunities for future research, particularly in terms of scalability. Expanding to larger datasets and more complex models could offer valuable insights, though it remains to be explored in future works.
\bibliography{main}

\appendix

\newtcolorbox{casebox}{
    colframe=black!50,    
    colback=white,       
    coltitle=black,      
    boxrule=0.4mm,       
    arc=1mm,             
    top=1mm,             
    bottom=1mm,          
    left=1.6mm,            
    right=1.6mm,           
    before skip=2mm,     
    after skip=2mm,      
}

\section{Appendix}

\subsection{Prompts for Baseline Models}
\label{sec:appendix_prompt_baseline}
We incorporate various prompt-based methods in our experiments to ensure a fair comparison. To maintain consistency, we keep most of the prompt content similar to that used for \themodel in Section~\ref{sec:ourprompt}. Below, we list the prompts used for the baseline models.

For the zero-shot and LoRA fine-tuning experiments, we use the following prompt:
\begin{mybox}
\textit{Based on the evidence, determine if the claim is supported by the evidence or refuted by it.\\
Claim: \texttt{[claim text $c$]}\\
Evidence: \texttt{(1)[evidence text $e_1$](2)...}}\\
\textit{Please respond with only whether the claim is ``Supported'' or ``Refuted.''}
\end{mybox}

For the zero-shot + CoT and STaR* experiments, the prompt is as follows:
\begin{mybox}
\textit{Based on the evidence, determine if the claim is supported by the evidence or refuted by it.\\
Claim: \texttt{[claim text $c$]}\\
Evidence: \texttt{(1)[evidence text $e_1$](2)...}}\\
\textit{Think step by step, output your response in the following format:\\
Chain: \texttt{[your reasoning chain]}\\
Answer:\texttt{[the claim is supported or the claim is refuted]}}
\end{mybox}

For the Few-shot experiment, the prompt is similar to the zero-shot prompt but includes examples:
\begin{mybox}
\textit{Based on the evidence, determine if the claim is supported by the evidence or refuted by it.\\
Please respond with only whether the claim is ``Supported'' or ``Refuted.'' Here are some examples:\\ 
Claim: Simon Grundel-Helmfelt is most ...\\
Evidence: (1) Baron Simon Grundel ... (2)...
Output: Refuted\\
\\(...more examples...)\\\\
Follow the above examples:\\
Claim: \texttt{[claim text $c$]}\\
Evidence: \texttt{(1)[evidence text $e_1$](2)...}\\
Output:
}
\end{mybox}

For the few-shot + structured CoT, we use the same prompt as \themodel, but with examples of structured reasoning:
\begin{mybox}
\textit{Based on the evidence, determine if the claim is supported by the evidence or refuted by it. Output the reasoning chain. Here are some examples:\\ 
Claim: Simon Grundel-Helmfelt is most ...\\
Evidence: (1) Baron Simon Grundel ... (2)...
Chain: C1: Simon Grundel ... \\
\\(...more examples...)\\\\
Follow the above examples:\\
Claim: \texttt{[claim text $c$]}\\
Evidence: \texttt{(1)[evidence text $e_1$](2)...}\\
Chain:
}
\end{mybox}

\subsection{Prompts with Hint}\label{sec:prompthint}
In \themodel, we add hint and regenerate reasoning chains for the ones that falsely predict the label of the claim. If the truth label is $p=\textit{Supported}$, we use prompt:
\begin{mybox}
\textit{Based on the evidence, determine if the claim is supported by the evidence or refuted by it. Output the reasoning chain.\\
Claim: \texttt{[claim text $c$]}\\
Evidence: \texttt{(1)[evidence text $e_1$](2)...}}\\
\textit{Hint: Every detail in this claim is supported.}
\end{mybox}
If the truth label is $p=\textit{Refuted}$, we use prompt:
\begin{mybox}
\textit{Based on the evidence, determine if the claim is supported by the evidence or refuted by it. Output the reasoning chain.\\
Claim: \texttt{[claim text $c$]}\\
Evidence: \texttt{(1)[evidence text $e_1$](2)...}}\\
\textit{Hint: The claim should be refuted, locate the error in the reasoning chain.}
\end{mybox}

\subsection{Dataset Statistics}\label{sec:datasetinfo}
The following table presents the information of the dataset (validation set) that we have tested on. ``HV'' represents HOVER, while ``FS'' stands for FEVEROUS.
\begin{table}[h]  
  \centering  
  \resizebox{\columnwidth}{!}{  
  \setlength{\tabcolsep}{3pt}
    \begin{tabular}{ccccc}  
    \toprule
    \textbf{Dataset} & \textbf{HV-2} & \textbf{HV-3} & \textbf{HV-4} & \textbf{FS-S} \\
    \midrule  
    Total  & 1,126 & 1,835 & 1,039 & 2,962 \\
    Supported Claims & 521 & 968 & 511 & 1411 \\
    Refuted Claims & 605 & 867 & 528 & 1551\\
    Avg. Words in Claim & 19.6 & 24.1 & 32.2 & 27.6 \\
    Avg. Evidence Pieces & 2.0 & 3.0 & 4.0 & 2.1 \\
    Avg. Words in Evidence & 137.3 & 211.1 & 278.2 & 50.3 \\
    \bottomrule 
    \end{tabular}%
    }  
  \label{tab:datasets}  
\end{table}

\subsection{\themodel-Generated Reasoning Chains}\label{sec:cases}
In this section, we present several reasoning chains generated by \themodel. The evidence is omitted due to its length, which makes it unsuitable for display. Red coloring is used to highlight where the reasoning chains become incorrect.
\small{\begin{casebox}
\textbf{Claim}: The artist whose work was displayed in 1974 at Debel Gallery, was closely associated with the Viennese Actionism group, while Howard Zieff was an American television commercial director.\\
\textbf{Label}: Supported\\\\
\textbf{Model Output}:\\
C1: The artist whose work was displayed in 1974 at Debel Gallery was closely associated with the Viennese Actionism group.\\Entity Resolution: Artist -> Rudolf Schwarzkogler (from E2)\\Resolution Verification: E1 confirms Rudolf Schwarzkogler's work was displayed in 1974 at Debel Gallery.\\Verification: E2 supports that Rudolf Schwarzkogler was closely associated with the Viennese Actionism group.\\Status: Supported.\vspace{1mm}\\C2: Howard Zieff was an American television commercial director.\\Entity Resolution: Howard Zieff -> Howard B. Zieff (from E3)\\Resolution Verification: E3 confirms Howard B. Zieff was an American director, television commercial director, and advertising photographer.\\Verification: E3 supports that Howard Zieff was an American television commercial director.\\Status: Supported.
\end{casebox}}

\small{\begin{casebox}
\textbf{Claim}: Ben Karlin wrote the 2013 episode of the TV show, Netflix, that was directed by the actor who played Kevin Arnold in ``The Wonder Years''\\
\textbf{Label}: Refuted\\\\
\textbf{Model Output}:\\
C1: Ben Karlin wrote the 2013 episode of the TV show, Netflix.\\Entity Resolution: TV show -> Modern Family (from E1)\\Resolution Verification: E1 confirms 'ClosetCon '13' aired in 2013, and E1 also confirms Ben Karlin wrote it.\\Verification: E1 supports that Ben Karlin wrote the 2013 episode, but it's clear that the TV show is Modern Family, not Netflix.\\Status: Refuted.\vspace{1mm}\\C2: The actor who played Kevin Arnold in \"The Wonder Years\" directed the 2013 episode.\\Entity Resolution: Actor -> Fred Savage (from E2)\\Resolution Verification: E2 confirms Fred Savage played Kevin Arnold in ``The Wonder Years''.\\Verification: E2 supports that Fred Savage played Kevin Arnold, but it's clear that he directed the episode in 2013, not that he played Kevin Arnold in 2013.\\Status: Refuted.\vspace{1mm}\\C3: The TV show was Netflix.\\Entity Resolution: TV show -> Modern Family (from E1)\\Resolution Verification: E1 confirms 'ClosetCon '13' aired on Modern Family, not Netflix.\\Verification: E1 refutes that the TV show was Netflix.\\Status: Refuted.
\end{casebox}}

\small{\begin{casebox}
\textbf{Claim}: The builder of Shadow Creek Golf Course is an Jewish real estate developer. The builder also owns the Encore hotel and casino in Las Vegas.\\
\textbf{Label}: Refuted\\\\
\textbf{Model Output}:\\
C1: The builder of Shadow Creek Golf Course is an Jewish real estate developer.\\Entity Resolution: Builder of Shadow Creek Golf Course -> Steve Wynn (from E3)\\Resolution Verification: E3 confirms Steve Wynn built Shadow Creek Golf Course.\\Verification: E3 shows Steve Wynn built Shadow Creek Golf Course, but it doesn't provide any information about his religion. There's no evidence to support the claim that Steve Wynn is Jewish.\\Status: Refuted.\vspace{1mm}\\C2: The builder also owns the Encore hotel and casino in Las Vegas.\\Entity Resolution: Builder -> Steve Wynn (from E3)\\Resolution Verification: E3 confirms Steve Wynn built Shadow Creek Golf Course, and E2 confirms Wynn Resorts owns Encore.\\Verification: E3 shows Steve Wynn built Shadow Creek Golf Course, and E2 confirms Wynn Resorts owns Encore.\\Status: Supported.
\end{casebox}}

\small{\begin{casebox}
\textbf{Claim}: Carnegie Mellon University, not the institution which is home to the Lyme Academy of Fine Arts, is a university in Pennsylvania.\\
\textbf{Label}: Supported\\\\
\textbf{Model Output}:\\
C1: Carnegie Mellon University, not the institution which is home to the Lyme Academy of Fine Arts, is a university in Pennsylvania.\\Entity Resolution: institution which is home to the Lyme Academy of Fine Arts -> University of New Haven (from E1)\\Resolution Verification: E1 confirms Lyme Academy College of Fine Arts is part of University of New Haven, not CMU. E3 confirms CMU is in Pennsylvania.\\Verification: E1 refutes \textcolor{red}{the institution which is home to the Lyme Academy of Fine Arts being CMU}. E3 supports CMU being in Pennsylvania.\\Status: Refuted.
\end{casebox}}
(The model misunderstood the meaning of the claim.)

\small{\begin{casebox}
\textbf{Claim}: Orange is the main flavor of both \textcolor{red}{zero-calorie} soft drinks, Solo from Norway, and the drink advertised with Krupa (song).\\
\textbf{Label}: Refuted\\\\
\textbf{Model Output}:\\
C1: Orange is the main flavor of Solo from Norway.\\Verification: E1 confirms Solo is an orange-flavored soft drink from Norway.\\Status: Supported.\vspace{1mm}\\C2: Orange is the main flavor of the drink advertised with Krupa (song).\\Entity Resolution: Drink advertised with Krupa -> Sunkist\\Resolution Verification: E3 confirms Krupa is from an Sunkist advert.\\Verification: E3 confirms Krupa is from an Sunkist advert and E2 confirms Sunkist is primarily orange flavored.\\Status: Supported.
\end{casebox}}
(The model lost the information of "zero-calorie" in the claim during the Claim Decomposition process, leading to an error.)

\small{\begin{casebox}
\textbf{Claim}: The star of Forever Strong played a character that is based on the Hanna-Barbera show ``Birdman and the Galaxy Trio''.\\
\textbf{Label}: Supported\\\\
\textbf{Model Output}:\\
C1: The star of Forever Strong played a character that is based on the Hanna-Barbara show ``Birdman and the Galaxy Trio''.\\Entity Resolution: Star of Forever Strong -> Gary Cole (from E1)\\Resolution Verification: E1 confirms Gary Cole was a star of Forever Strong.\\Entity Resolution: \textcolor{red}{Character played by Gary Cole -> ? (from E2)}\\Resolution Verification: E2 confirms Gary Cole voiced Harvey Birdman, but it doesn't mention him playing \textcolor{red}{a character in Forever Strong}.\\Verification: E1 shows Gary Cole was a star of Forever Strong, but E2 shows \textcolor{red}{he voiced Harvey Birdman, not a character in Forever Strong}. Therefore, the claim is refuted.\\Status: Refuted.
\end{casebox}}
(The model incorrectly associated the character in the claim with ``Forever Strong,'' leading to an error in Entity Resolution.)
\end{document}